\documentclass[10pt,twocolumn,letterpaper]{article}

\usepackage{cvpr}
\usepackage{times}
\usepackage{epsfig}
\usepackage{graphicx}
\usepackage{amsmath}
\usepackage{amssymb}

\usepackage{amsmath,amsfonts,bm}

\def\eqref#1{equation~\ref{#1}}

\def\1{\bm{1}}

\def\vc{{\bm{c}}}

\def\vh{{\bm{h}}}

\def\mW{{\bm{W}}}

\DeclareMathAlphabet{\mathsfit}{\encodingdefault}{\sfdefault}{m}{sl}
\SetMathAlphabet{\mathsfit}{bold}{\encodingdefault}{\sfdefault}{bx}{n}

\usepackage{makecell}
\usepackage{multirow}
\usepackage{algorithm}
\usepackage{algorithmic}
\usepackage{caption}
\usepackage{tablefootnote}

\cvprfinalcopy %

\def\cvprPaperID{3587} %

\usepackage{xargs} 
\usepackage{xcolor}  %
\usepackage[colorinlistoftodos,prependcaption,textsize=tiny]{todonotes}
\newcommandx{\unsure}[2][1=]{\todo[linecolor=red,backgroundcolor=red!25,bordercolor=red,#1]{#2}}
\newcommandx{\tochange}[2][1=]{\todo[linecolor=green,backgroundcolor=green!25,bordercolor=green,#1]{#2}}
\newcommandx{\info}[2][1=]{\todo[linecolor=yellow,backgroundcolor=yellow!25,bordercolor=yellow,#1]{#2}}
\newcommandx{\changed}[2][1=]{\todo[linecolor=blue,backgroundcolor=blue!25,bordercolor=blue,#1]{#2}}
\newcommandx{\thiswillnotshow}[2][1=]{\todo[disable,#1]{#2}}

\newcount\Comments
\usepackage{color}
\definecolor{darkgreen}{rgb}{0,0.6,0}
\definecolor{orange}{rgb}{1,0.5,0}
\newcommand{\kibitz}[2]{\ifnum\Comments=0{\color{#1}{#2}}\fi}

\definecolor{citecolor}{RGB}{34, 139, 34}
\usepackage[pagebackref=true,breaklinks=true,colorlinks,bookmarks=false, citecolor=citecolor]{hyperref}

\begin{document}

\title{The Regretful Agent: Heuristic-Aided Navigation through Progress Estimation}

\newcommand\blfootnote[1]{%
  \begingroup
  \renewcommand\thefootnote{}\footnote{#1}%
  \addtocounter{footnote}{-1}%
  \endgroup
}

\newcommand*\samethanks[1][\value{footnote}]{\footnotemark[#1]}

\author{Chih-Yao~Ma\textsuperscript{$* \dagger$}, Zuxuan~Wu\textsuperscript{$\ddagger$}, Ghassan~AlRegib\textsuperscript{$\dagger$}, Caiming~Xiong\textsuperscript{$\S$}, Zsolt~Kira\textsuperscript{$\dagger$} \\
\normalsize
\textsuperscript{$\dagger$}Georgia Institute of Technology,
\textsuperscript{$\ddagger$}University of Maryland, College Park,
\textsuperscript{$\S$}Salesforce Research
}

\twocolumn[{
\renewcommand\twocolumn[1][]{#1}
\maketitle
\vspace*{-0.5cm}
\centering
\captionsetup{type=figure}
\includegraphics[width=1\linewidth]{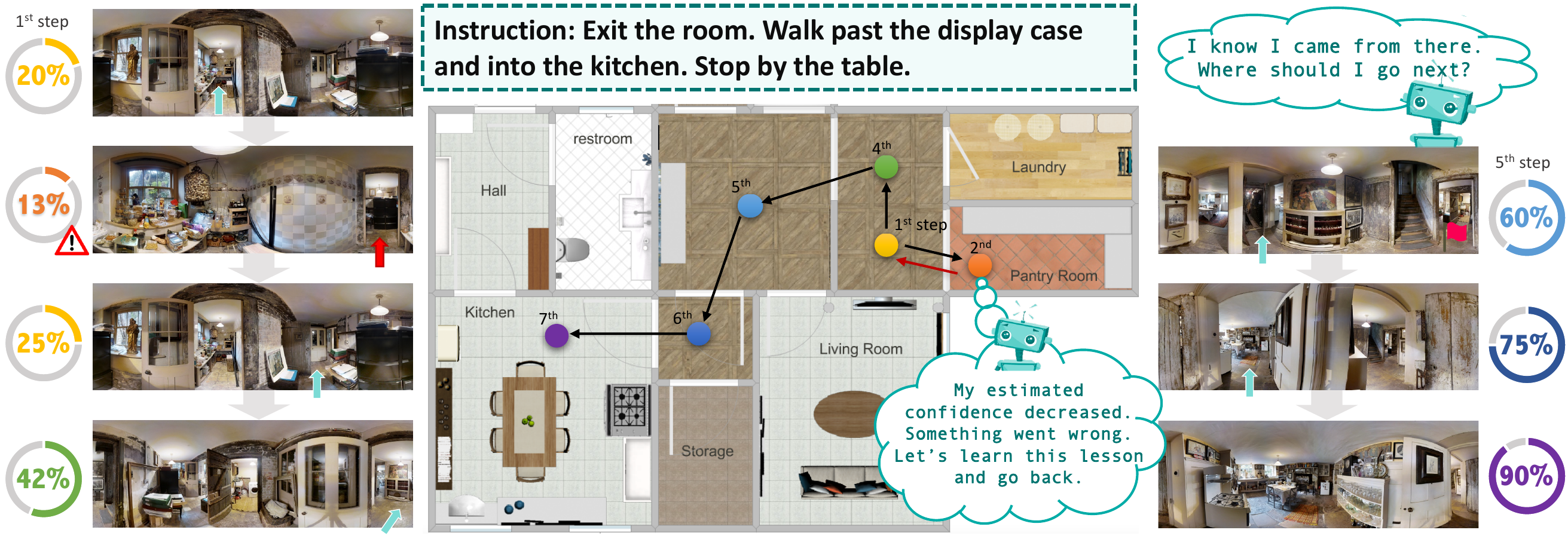}
\vspace{-0.5cm}
\captionof{figure}{
  Vision-and-Language Navigation task and our proposed regretful navigation agent. 
  The agent leverages the self-monitoring mechanism~\cite{ma2019selfmonitoring} through time to decide when to roll back to a previous location and resume the instruction-following task. 
  Our code is available at \href{https://github.com/chihyaoma/regretful-agent}{https://github.com/chihyaoma/regretful-agent}.
  }
\label{fig:concept}
\vspace*{0.5cm}
}]

\maketitle

\thispagestyle{plain}

\begin{abstract}
\blfootnote{$^{*}$ Work partially done while the author was a research intern at Salesforce Research.}
  As deep learning continues to make progress for challenging perception tasks, there is increased interest in combining vision, language, and decision-making.
  Specifically, the Vision and Language Navigation (VLN) task involves navigating to a goal purely from language instructions and visual information without explicit knowledge of the goal. 
  Recent successful approaches have made in-roads in achieving good success rates for this task but rely on beam search, which thoroughly explores a large number of trajectories and is unrealistic for applications such as robotics. 
  In this paper, inspired by the intuition of viewing the problem as search on a navigation graph, we propose to use a progress monitor developed in prior work as a learnable heuristic for search. 
  We then propose two modules incorporated into an end-to-end architecture: 1) A learned mechanism to perform backtracking, which decides whether to continue moving forward or roll back to a previous state (Regret Module) and 2) A mechanism to help the agent decide which direction to go next by showing directions that are visited and their associated progress estimate (Progress Marker). 
  Combined, the proposed approach significantly outperforms current state-of-the-art methods using greedy action selection, with 5\% absolute improvement on the test server in success rates, and more importantly 8\% on success rates normalized by the path length.  
\end{abstract}

\vspace{-10pt}
\section{Introduction}

Building on the success of deep learning in solving various computer vision tasks, several new tasks and corresponding benchmarks have been proposed to combine visual perception and decision-making~\cite{anderson2018vision, wu2018building, das2018embodied, ai2thor, savva2017minos, xia2018gibson, brodeur2017home}. One such task is the Vision-and-Language Navigation task (VLN), where an agent must navigate to a goal purely from language instructions and visual input without explicit knowledge of the goal. This task has a number of applications, including service robotics where it would be preferable if humans interacted naturally with the robot by instructing it to perform various tasks. 

Recently, there have been several approaches proposed to solve this task. The dominant approaches frame the navigation task as a {\it sequence to sequence} problem~\cite{anderson2018vision}. Several enhancements such as synthetic data augmentation~\cite{fried2018speaker}, pragmatic inference~\cite{fried2018speaker}, and combinations of model-free and model-based reinforcement learning techniques~\cite{wang2018look} have also been proposed. 
However, current methods are separated into two regimes: those that use beam search and obtain good success rate (with longer trajectory lengths) and those that use greedy action selection (and hence result in very low trajectory lengths) but obtain much lower success rates. In fact, there have recently been new metrics proposed that balance these two objectives~\cite{anderson2018evaluation}. Intuitively, the agent should perform intelligent action selection (akin to best-first search), without exhaustively exploring the search space. For robotics application, for example, the use of beam search is unrealistic as it would require the robot to explore a large number of possible trajectories. 
 
In this paper, we view the process of navigation as {\it graph search} across the navigation graph and employ two strategies, encoded within the neural network architecture, to enable navigation without the use of beam search. Specifically, we develop: 1) A {\bf Regret Module} that provides a mechanism to allow the agent to {\it learn when to backtrack}~\cite{fikes1972learning, barraquand1991robot} and 2) We propose a {\bf Progress Marker} mechanism that allows the agent to incorporate information from previous visits and reason about such visits and their associated progress estimates towards better action selection. 
 
Specifically, in graph search a heuristic is used to make meaningful progress towards the goal in a manner that avoids exhaustive search but is more effective than na\"ive greedy search. We therefore build on recent work~\cite{ma2019selfmonitoring} that developed a {\it progress monitor} which is a learned mechanism that was used to estimate the progress made towards the goal (with low values meaning progress has not been made and high values meaning the agent is closer to the goal). In that work, however, the focus was on the regularizing effect of the progress monitor as well as its use in beam search. Instead, we use this progress monitor effectively as a {\it learned heuristic} that can be used to determine directions that are more likely to lead towards the goal during inference. 

We use the Progress Marker in two ways. First, we leverage the notion of backtracking, which is prevalent in graph search, by developing a {\it learned rollback mechanism} that decides whether to go back to the previous location or not (Regret Module). Second, we incorporate a mechanism to allow the agent to use the estimated progress it computed when visiting the viewpoints to choose the next action to perform after it has rolled back (Progress Marker). This allows the agent to know when particular directions have already been visited and the progress they resulted in, which can bias it to not re-visit states unless warranted.
We do this by augmenting the visual state vectors with the progress estimates so that the agent can reduce the probability of revisiting such states (again, in a learned manner). 

We demonstrate that these learned mechanisms are superior to greedy decoding. Our agent is able to achieve state-of-the-art results among published works both in terms of success rate (when beam search is not used) and more importantly the SPL~\cite{anderson2018evaluation} metric which incorporates path length, owing to our short trajectory lengths. In summary, our contributions include: 1) A graph search perspective on the instruction-based navigation problem, and use of a {\bf learned heuristic} in the form of a progress monitor to effectively explore the navigation graph, 2) an end-to-end trainable {\bf Regret Module} that can learn to decide when to roll back to the previous location given the history of textual and visual grounding observed, 3) a {\bf Progress Marker} that can enable effective backtracking and reduce the probability of going to a visited location accordingly, and 4) state-of-the-art results on the VLN task.

\section{Related Work}

{\bf Vision and language navigation.} There are a number of benchmarks and environments for investigating the combination of vision, language, and decision-making. This includes House3D~\cite{wu2018building}, Embodied QA~\cite{das2018embodied}, AI2-THOR~\cite{ai2thor}, navigation based agents~\cite{mousavian2018visual, wayne2018unsupervised, mirowski2017learning} (including with communication~\cite{de2018talk}), and the VLN task that we focus on~\cite{anderson2018vision}. For tasks that contain only sparse rewards, reinforcement learning approaches exist~\cite{wang2018look, yu2018guided, das2018neural}, for example focusing on language grounding through guided feature transformation~\cite{yu2018guided}  and development of a neural module approach~\cite{das2018neural}. Our work, in contrast, focuses on tasks that contain language instructions that can guide the navigation process and has applications such as service robotics. Approaches to this task are dominated by a sequence-to-sequence formulation, beginning with initial work introducing the task~\cite{anderson2018vision}. Subsequent methods have used a Speaker-Follower technique to generate synthetic instructions that are used for data augmentation and pragmatic inference~\cite{fried2018speaker}, as well as the combination of supervised-based and RL-based approaches~\cite{wang2018look, wang2019reinforced}. Recently, the Self-Monitoring navigation agent was introduced which learns to estimate progress made towards the goal using visual and language co-grounding~\cite{ma2019selfmonitoring}. 
Prior work employs beam-search type techniques, though, optimizing for success rate at the expense of trajectory length and reduced applicability to robotics and other domains. Inspired by the latter work, we view the progress monitor as a learned heuristic and combine it with other techniques in graph search, namely backtracking, to use it for {\it action selection}, which was not a focus of the prior work.

{\bf Navigation and learned heuristics.} Several works in vision and robotics have explored the intersection of learning and planning. In robotics, planning systems must often explore large search trees for getting from start to goal, and selection of the next state to expand must be done intelligently to reduce computation. Often fixed heuristics (e.g. distance to goal) are used, but these are static, require known goal locations, and are used for optimal A*-style algorithms rather than greedy best-first search, which is what can be employed on robots when maps are not available~\cite{russell2016artificial}. Recently, several learning-based approaches have been developed for such heuristics, including older works that learn residuals for existing heuristics~\cite{xu2007discriminative}, heuristic ranking methods that enable refinement of new ones~\cite{wilt2015building} as well as learning of a heuristic policy in a Markov Decision Process (MDP) formulation to directly optimize search effort by taking into account history and contextual information~\cite{bhardwaj2017learning}. In our work, we similarly learn to estimate a heuristic (progress monitor) and use it for action selection, showing that the resulting estimates can generalize to unseen environments. We also develop an architecture to explicitly learn when to backtrack based on this progress monitor (with a Progress Marker to reduce the chance of choosing the same action again after backtracking unless warranted), which further improves navigation performance.

\textbf{Modern Reinforcement Learning.}
Modern Reinforcement Learning methods like Asynchronous Advantage Actor Critic (A3C)~\cite{mnih2016asynchronous} or Advantage Actor Critic (A2C) methods are related to the baseline Self-Monitoring agent~\cite{ma2019selfmonitoring} and the proposed Regretful agent.
Specifically, the progress monitor in the Self-Monitoring agent (our baseline) is similar to the value function in RL, and the difference between progress marker of a viewpoint and current progress estimation (denote as $\Delta \bm{v}^{marker}_{t,k}$, see Sec.~\ref{sec:progress_marker}) is conceptually similar to the advantage function.
However, the advantage function in RL serves as a way to regularize and improve the training of the policy network. 
We instead associate the $\Delta \bm{v}^{marker}_{t,k}$ directly to all navigable states, and this $\Delta \bm{v}^{marker}_{t,k}$ has a direct impact on the agent deciding next action even during inference.
While having an accurate value estimate for VLN with dynamic and implicit goals may reduce the need for this formulation, we however believe that this is hardly possible because of the lack of training data.
On the other hand, relating to the proposed end-to-end learned regret module,
\textit{Leave no Trace}~\cite{eysenbach2018leave}
learns a forward and a \textit{reset} policy to reset the environment for preventing the policy entering a non-reversible state.
Instead of learning to reset, we learn to rollback to a previous state and continue the navigation task with a policy network that learns to decide a better next step.

\section{Baseline}
\label{sec:notation}
Given natural language instructions, our task is to train an agent to follow these instructions and reach an  (unspecified) goal in the environment (see Figure~\ref{fig:concept} for an example). This requires processing both the instructions and the visual inputs, along with attentional mechanisms to ground them to the current situation. We adapt the recently introduced Self-Monitoring Visual-Textual Co-grounding agent~\cite{ma2019selfmonitoring} as our baseline.
The Self-Monitoring agent consists of two primary components: 
(1) A visual-textual co-grounding module that grounds to the completed instruction, the next instruction, and the subsequent navigable directions represented as visual features.
(2) A progress monitor that takes the attention weights of grounded instructions as input and estimates the agent's progress towards completing the instruction. It was shown that such a progress monitor can {\it regularize} the attentional mechanism (via an additional loss), but the authors did not focus on using the progress estimates for action selection itself. In the following, we briefly introduce the Self-Monitoring agent.

Specifically, a language instruction with $L$ words is represented via embeddings denoted as
$\bm{X}=\big\{\bm{x}_1, \bm{x}_2, \ldots, \bm{x}_L\big\}$, where $\bm{x}_l$ is the feature vector for the $l$-th word encoded by a Long Short-Term Memory (LSTM) language encoder. 
Following~\cite{ma2019selfmonitoring,fried2018speaker}, we use a panoramic view as visual input. At the $t$-th time step, the agent perceives a set of images at each viewpoint 
$\bm{v}_t = \big\{\bm{v}_{t,1}, \bm{v}_{t,2}, ..., \bm{v}_{t,K}\big\},$ where $K$ is the maximum number of navigable directions, and $\bm{v}_{t,k}$ represents the image feature of direction $k$ obtained from an ImageNet pre-trained ResNet-152.
It first obtains visual and textual grounded features, $\hat{\bm{v}}_t$, and $\hat{\bm{x}}_t$, respectively, with hidden states from the last time step $\bm{h}_{t-1}$ using soft-attention (see~\cite{ma2019selfmonitoring} for details). Conditioned on these grounded features and historical context, it then produces the hidden context of the current step $\bm{h}_t$: 
\begin{equation*}
\label{eq:lstm}
\bm{h}_t, \bm{c}_t = LSTM([\hat{\bm{x}}_{t}, \hat{\bm{v}}_{t}, \mathcal{\bf a}_{t-1}], \bm{h}_{t-1}, \bm{c}_{t-1}) ,
\end{equation*}
where $[,]$ denotes concatenation and $\bm{c}_{t-1}$ denote cell states from the last time step.
To further decide where to go next, the current hidden states $\bm{h}_t$ are concatenated with grounded instructions $\hat{\bm{x}}_{t}$, yielding a representation that contains historical context and relevant parts of the instructions (for example, corresponding to parts that have just been carried out and those that have to be carried out next), to compute the correlations with visual features for each viewpoint $k$ ($\bm{v}_{t, k}$). Formally, action selection is calculated as follows:
\begin{equation*}
\label{eq:action-selection}
o_{t,k} = (\bm{W}_{a} [\bm{h}_t, \hat{\bm{x}}_{t}])^\top g(\bm{v}_{t, k})
\quad\text{and}\quad 
\bm{p}_{t} =  \textrm{softmax}(\bm{o}_t)
\end{equation*}
where $\bm{W}_a$ are the learned parameters and $g(\cdot)$ is a Multi-Layer Perceptron (MLP).

Furthermore, we also equip the agent with a progress monitor following~\cite{ma2019selfmonitoring} to enforce the attention weights of the textual grounding to align with the progress made toward the goal, further regularizing the grounded instructions to be relevant. 
The progress monitor is optimized such that the agent is required to use the attention distribution of textual grounding to predict the distance from goal. 
The output of progress monitor $p^{pm}_t$ represents the completeness of instruction-following estimated by the agent. 
\begin{equation*}
\label{eq:progress-monitor}
\vh^{pm}_{t} = \sigma(\mW_{h} ([\vh_{t-1}, \hat{\bm{v}}_t]) \otimes tanh(\vc_{t}))
\end{equation*}
\begin{equation*}
p^{pm}_{t} = tanh(\mW_{pm}([\pmb{\alpha}_{t}, \vh^{pm}_{t}]))
\end{equation*}
where 
$\bm{W}_h$ and $\bm{W}_{pm}$ are the learnt parameters,
$c_t$ is the cell state of the LSTM,
$\otimes$ denotes the element-wise product,
$\bm{\alpha}_t$ is the attention weights of textual grounding,
and $\sigma$ is the sigmoid function.
Please refer to \cite{ma2019selfmonitoring} for further details on the baseline architecture.

\section{Regretful Navigation Agent}
\begin{figure}[t]
\begin{center}
\includegraphics[width=1.0\linewidth]{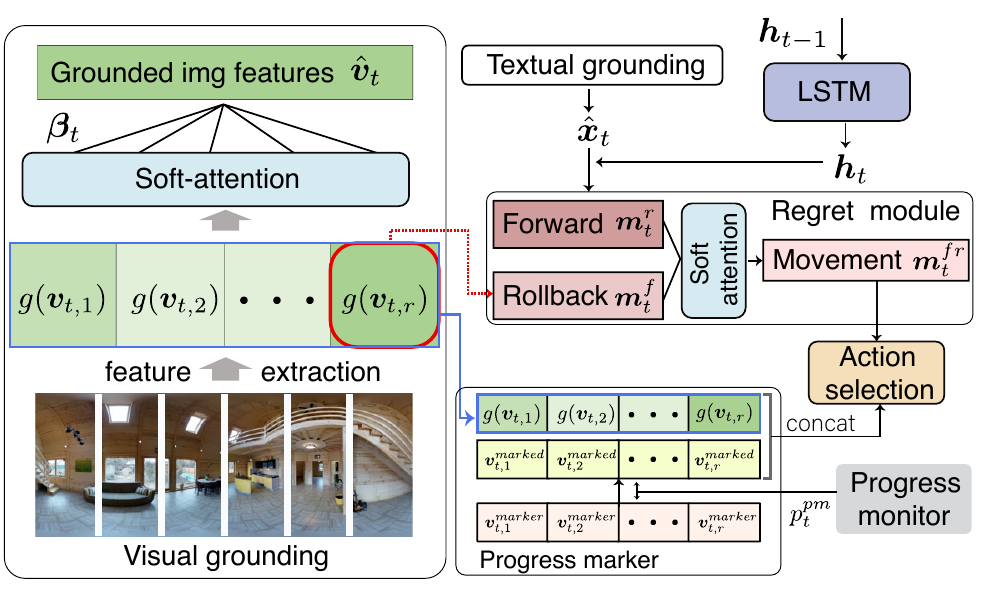}
\end{center}
   \caption{
   Illustration of the proposed regretful navigation agent.
   Note that the progress monitor is based on \cite{ma2019selfmonitoring}.
   }
\label{fig:framework}
\vspace{-0.1in}
\end{figure}

The progress monitor previously mentioned reflects the agent's progress made towards the goal, and consequently its outputs will decrease or fluctuate if the agent selects an action leading to deviation from the goal. Conversely it will increase if it moves closer to the goal by completing the instruction. We posit that such a property, while conceptually simple, provides critical feedback for action selection. To this end, we leverage the outputs of the progress monitor to allow the agent to regret and backtrack using a \textbf{Regret Module} and a \textbf{Progress Marker} (see Figure.~\ref{fig:framework}). In particular, the Regret Module examines the progress made from the last step to the current step to decide whether to take a \textit{forward} or \textit{rollback} action. Once the agent regrets and rolls back to the previous location, the Progress Marker informs whether location(s) have been visited before and rates the visited location(s) according to the agent's confidence in completing the instruction-following task. Combining the two proposed methods, we show that the agent is able to perform a local search on the navigational graph by (1) assessing the current progress, (2) deciding when to roll back, and (3) selecting the next location after rollback occurs. In the following, we elaborate these two components in detail.

\subsection{Regret Module}
The {\bf Regret Module} takes in the outputs of the progress monitor at different time steps and decides whether to go \textit{forward} or to \textit{rollback}. In particular, we use the concatenation of hidden state $\bm{h}_t$ and grounded instruction $\hat{\bm{x}}_{t}$ as our \textit{forward} embedding $\bm{m}^{f}_{t}$, and more importantly we introduce a \textit{rollback} embedding $\bm{m}^{r}_{t}$ to be the projection of the visual features for the action that leads to the previously visited location. The two vector representations are as follows:
\begin{equation*}
\bm{m}^{f}_{t} = \bm{W}_{a} [\bm{h}_t, \hat{\bm{x}}_{t}]
\quad\text{and}\quad 
\bm{m}^{r}_{t} = g(\bm{v}_{t, r}) ,
\end{equation*}
where $\bm{W}_a$ are the learned parameters, 
$\hat{\bm{x}}_{t}$ is the grounded instruction obtained from the textual grounding module, and 
$\bm{v}_{t, r}$ is the image feature vector representing a direction that points to the previously visited location.

To decide whether to go forward or rollback, the Regret Module leverages the difference of the progress monitor outputs between the current time step and the previous time step $\Delta p^{pm}_{t} = p^{pm}_{t} - p^{pm}_{t-1}$.
Intuitively, if the difference is larger than a certain threshold $\Delta p^{pm}_{t} > \sigma$, the agent should decide to take a \textit{forward} action, and vice versa. Since it is hard to decide an optimal value for $\sigma$, we achieve this by computing attention weights $\bm{\alpha}^{fr}_t$ and perform a weighted sum on both \textit{forward} and \textit{rollback} embeddings. If the weight on \textit{rollback} is larger, the agent is likely to be biased to take an action that leads to the last visited location. Formally, the weights can be computed as:

\begin{equation*}
\bm{\alpha}^{fr}_t = \textrm{softmax}(\bm{W}_{r} (\Delta p^{pm}_{t}))
\end{equation*}
\begin{equation*}
\bm{m}^{fr}_{t} = (\bm{\alpha}^{fr}_t)^\top [\bm{m}^f_t, \bm{m}^r_t] ,
\end{equation*}
where $\bm{W}_r$ are the learnt parameters, 
$[,]$ denotes concatenation between feature vectors,
and $\bm{m}^{fr}_{t}$ represents the weighted sum of the \textit{forward} and \textit{rollback} embeddings. 
Note that to ensure the progress monitor remains focused on estimating the agent's progress and regularizing the textual grounding module, we \textit{detach} the output of the progress monitor which is fed into the Regret Module and set it as a leaf in the computational graph. 

\textbf{Action selection.}
Similar to existing work, the agent determines which image features from navigable directions have the highest correlation with the movement vector $\bm{m}^{fr}_{t}$ by computing the inner-product, and the probability of each navigable direction is then computed as:
\begin{equation*}
\bm{o}_{t,k} = (\mW_{fr} \bm{m}^{fr}_{t})^\top g(\bm{v}_{t, k}) 
\quad\text{and}\quad 
\bm{p}_{t} =  \textrm{softmax}(\bm{o}_t) ,
\end{equation*}
where $\mW_{fr}$ are the learned parameters and $\bm{p}_t$ is the probability distribution over navigable directions at time $t$.
In practice, once the agent takes a rollback action, we block the action that leads to oscillation.

\subsection{Progress Marker}
\label{sec:progress_marker}
The Regret Module provides a mechanism for the agent to decide when to rollback to a previous location or move forward according to the progress monitor outputs.
Once the agent rolls back, it is required to select the next direction to go forward. 
It is thus essential for the agent to (1) know which directions it has already visited (and rolled back) and (2) estimate if the visited locations can lead to a path which completes the given instruction. 

\begin{figure}[t]
\begin{center}
    \includegraphics[width=0.45\textwidth]{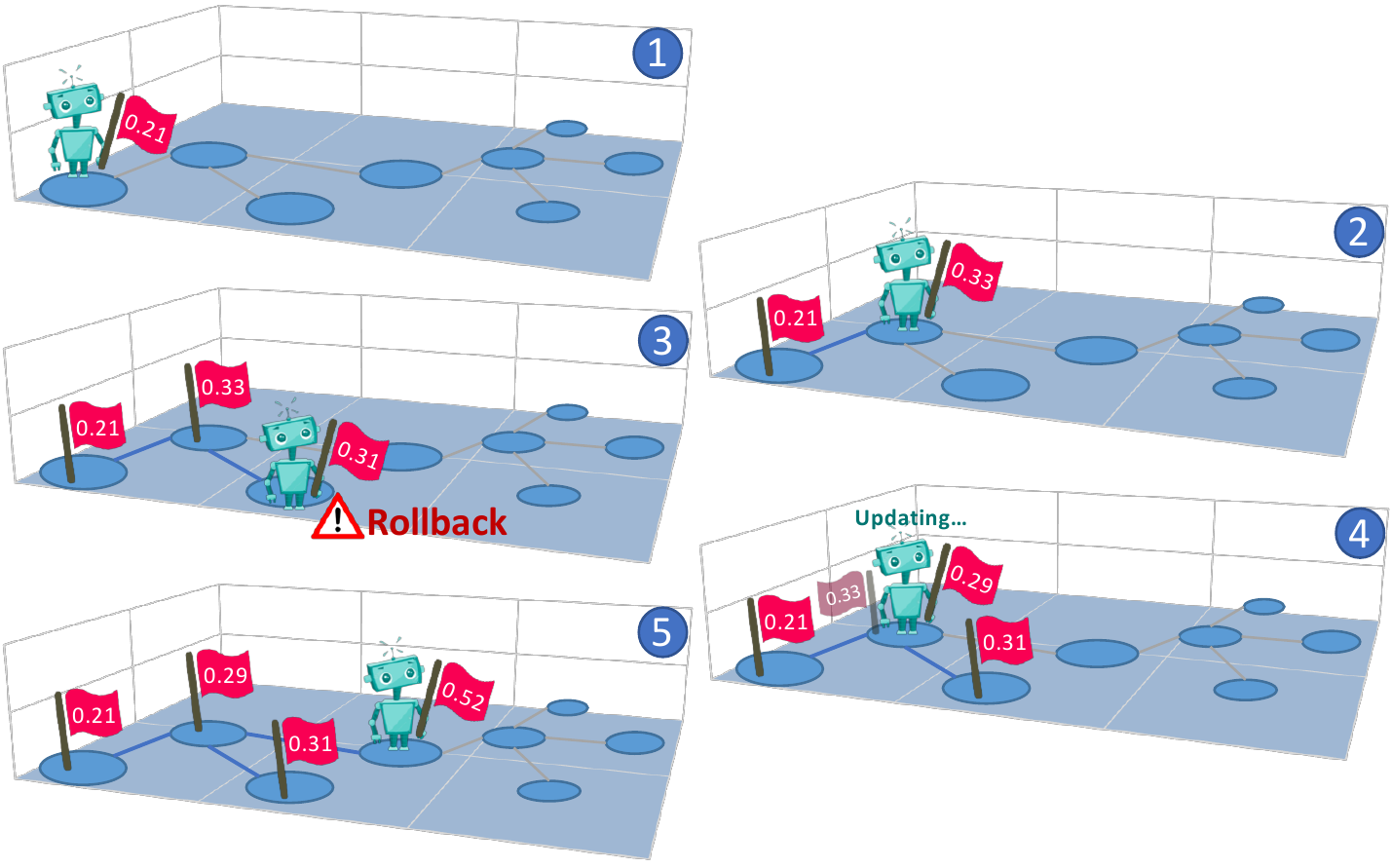}
\end{center}
\caption{
Concept of the proposed Progress Marker (red flags).
The agent marks each visited location with estimated progress made towards the goal. The changes on the estimated progress determines whether the agent should \textit{rollback} or \textit{forward}, and the difference between the current estimated progress and the markers on the next navigable directions helps the agent decide which direction to go.
}
\label{fig:marker_concept}
\vspace{-0.1in}
\end{figure}

Toward this end, we propose the \textbf{Progress Marker} to mark each visited location with the agent's confidence in completing the instruction (see Figure~\ref{fig:marker_concept}).
More specifically, we maintain a set of memory $\bm{M}^{}$ and store the output of the progress monitor associated with each visited location; if the location is not yet visited, the marker will be filled with a value 1: 
\[
  \bm{v}^{marker}_{t,k}=\begin{cases}
               p^{pm}_{i}, & \text{if $k$ leads to a location $i \in \bm{M}^{}$}.\\
               1, & \text{otherwise}.\\
            \end{cases}
\]
where $i$ is a unique viewpoint ID for each location.
We allow the marker on each location to be updated every time the agent visits it.

The marker value on each navigable direction indicates the estimated confidence that a location leads to the goal. 
We assign a value 1 for unvisited directions to encourage the agent to explore the environment.
The navigating probabilities between unvisited directions depend on the action probabilities $\bm{p}_t$ since their marker values are the same.

\textbf{Action selection with Progress Marker.}
During action selection, 
in addition to the movement vector $\bm{m}^{fr}_t$ that the agent can rely on in deciding which direction to go, 
we propose to label the marker value to each navigation direction as indications of whether a direction is likely to lead to the goal or to unexplored (and potentially better) paths. To achieve this, we leverage the difference between the current estimated progress and the marker for each navigable direction $\Delta \bm{v}^{marker}_{t,k} = p^{pm}_t - \bm{v}^{marker}_{t,k}$.
We then concatenate it to the visual feature representation for each navigable direction before action selection. 
\begin{equation*}
\bm{v}^{marked}_{t,k} = [g(\bm{v}_{t,k}), \Delta \bm{v}^{marker}_{t,k}] .
\end{equation*}
The difference $\Delta \bm{v}^{marker}_{t,k}$ indicates the chances of navigable directions leading to the goal and further inform the agent which direction to select.
In our design, lower $\Delta \bm{v}^{marker}_{t,k}$ corresponds to higher chance for action selection. 
For instance, in step 4 in Figure~\ref{fig:marker_concept}, the $\Delta \bm{v}^{marker}_{t,k}$ for starting location and the last visited location are 0.08 and -0.02 respectively, whereas an unvisited location will have -0.71, which eventually leads to 0.52 estimated progress.

When using Progress Marker, the final action selection is formulated as:
\begin{equation*}
\bm{o}_{t,k} = (\mW_{fr}\bm{m}^{fr}_{t})^\top \bm{v}^{marked}_{t,k}
\quad\text{and}\quad 
\bm{p}_{t} =  \textrm{softmax}(\bm{o}_t)
\end{equation*}
In practice, we tiled the difference $n$ times before concatenating with the projected image feature $\bm{v}_{t,k}$ in order to account for imbalance.
The marker value for the \textit{stop} action is set to be 0.

\subsection{Training and Inference}
We train the proposed agent with cross-entropy loss for action selection and Mean Squared Error (MSE) loss for progress monitor. 
In addition to these losses, we also introduce an additional entropy loss to encourage the agent to explore other actions, such that it is not biased to actions with already very high confidence. 
The motivation is that, after training an agent for a period of time, the agent starts to overfit and perform fairly well on the training set. 
As a result, the agent will not learn to properly roll back during training since the majority of the training samples do not require the agent to roll back. 
Introducing the entropy loss increases the chance of exploration and making incorrect actions during training.
\begin{align*}
\mathcal{L}_{loss} &= 
\lambda \overbrace{\sum_{t=1}^{T} y^{nv}_{t} log(p_{t,k})}^\text{action selection}
+ (1-\lambda)
\overbrace{\sum_{t=1}^{T} (y^{pm}_{t} - p^{pm}_{t})^2}^\text{progress monitor}\\
&\quad - \beta \underbrace{\sum_{t=1}^{T} \sum_{k=1}^{K} - p_{t,k} log(p_{t,k})}_\text{entropy loss} ,
\end{align*}
where 
$p_{t,k}$ is the action probability of each navigable direction,
$y^{nv}_{t}$ is the ground-truth navigable direction at step $t$,
$\lambda=0.5$ is the weight balancing the cross-entropy loss and MSE loss,
and $\beta=0.01$ is the weight for entropy loss.

Following existing approaches~\cite{ma2019selfmonitoring,fried2018speaker,anderson2018vision}, we perform categorical sampling during training for action selection.
During inference, the agent greedily selects the action with highest action probability.
\begin{table*}[t]
\centering
\small
\renewcommand{\arraystretch}{1.2}
\caption{
    Comparison with the state of the arts with greedy decoding for action selections\protect\footnotemark.
    *: with data augmentation.
}
\label{table:SOTAs}
\begin{tabular}{ccccccccccccc}
\Xhline{2\arrayrulewidth}
                 & \multicolumn{4}{c}{Validation-Seen} & \multicolumn{4}{c}{Validation-Unseen} & \multicolumn{4}{c}{Test (unseen)} \\ \cline{2-13} 
Method           & NE $\downarrow$         & SR $\uparrow$ & OSR $\uparrow$ & SPL $\uparrow$     & NE $\downarrow$          & SR $\uparrow$ & OSR  $\uparrow$   & SPL $\uparrow$    & NE $\downarrow$        & SR $\uparrow$        & OSR $\uparrow$ & SPL $\uparrow$      \\ \hline
Random           & 9.45       & 0.16       & 0.21  & -    & 9.23        & 0.16       & 0.22   & -    & 9.77      & 0.13      & 0.18   & 0.12   \\
Student-forcing~\cite{anderson2018vision}  & 6.01       & 0.39       & 0.53 & -     & 7.81        & 0.22       & 0.28   & -    & 7.85      & 0.20      & 0.27      & 0.18 \\
RPA~\cite{wang2018look}              & 5.56       & 0.43       & 0.53   & -   & 7.65        & 0.25       & 0.32  & -     & 7.53      & 0.25      & 0.33   & 0.23   \\
\makecell{Speaker-Follower~\cite{fried2018speaker}*} & 3.36       & 0.66       & 0.74   & -   & 6.62        & 0.36       & 0.45   & -    & 6.62      & 0.35      & 0.44   & 0.28   \\ 
\makecell{RCM~\cite{wang2019reinforced}*} & 3.37       & 0.67       & 0.77   & -   & 5.88        & 0.43       & 0.52   & -    & 6.01 & 0.43  & 0.51 & 0.35\\ 
\makecell{Self-Monitoring~\cite{ma2019selfmonitoring}*} & \textbf{3.22}       & 0.67       & \textbf{0.78}   & 0.58   & 5.52        & 0.45       & 0.56   & 0.32    & 5.99 & 0.43  & 0.55 & 0.32\\ 
\hline
Regretful       & 3.69      & 0.65       & 0.72       & 0.59      & 5.36       & 0.48       & \textbf{0.61}   & 0.37    & - & -  & - & - \\
Regretful*      & 3.23       & \textbf{0.69}       & 0.77       & \textbf{0.63}      & \textbf{5.32}        & \textbf{0.50}       & 0.59   & \textbf{0.41}    & \textbf{5.69} & \textbf{0.48}  & \textbf{0.56} & \textbf{0.40}\\
\Xhline{2\arrayrulewidth}          
\end{tabular}
\end{table*}
\footnotetext{Note that both Speaker-Follower~\cite{fried2018speaker} and Self-Monitoring~\cite{ma2019selfmonitoring} were originally designed to optimize the success rate (SR) via beam search, and concurrently to our work, RCM~\cite{wang2019reinforced} proposed a new setting allowing the agent to explore unseen environments prior to the navigation task via Self-Supervised Imitation Learning (SIL).}

\begin{table*}[t]
\centering
\small
\renewcommand{\arraystretch}{1.2}
\caption{
    Ablation study showing the effect of each proposed components compared to the prior arts.
    All methods here trained without data augmentation.
}
\label{table:ablation}
\begin{tabular}{cccccccccccc}
\Xhline{2\arrayrulewidth}
& & Regret & Progress &  \multicolumn{4}{c}{Validation-Seen} & \multicolumn{4}{c}{Validation-Unseen} \\ \cline{5-12} 
Method &\# & Module  & Marker  & NE $\downarrow$         & SR $\uparrow$ & OSR $\uparrow$ & SPL  $\uparrow$      & NE $\downarrow$          & SR $\uparrow$ & OSR $\uparrow$    & SPL  $\uparrow$       \\ \hline
Speaker-Follower~\cite{fried2018speaker} & & & & 4.86       & 0.52       & 0.63   & -   & 7.07        & 0.31       & 0.41  & -\\
\makecell{Self-Monitoring~\cite{ma2019selfmonitoring}} & & & & 3.72       & 0.63       & \textbf{0.75}   & 0.56   & 5.98        & 0.44       & 0.58    & 0.30       \\ 
\hline 

\multirow{3}{*}{Regretful} & 1  & \checkmark    &       & 3.88 & 0.64 & 0.70 & 0.58 & 5.65 & 0.47 & 0.59 & \textbf{0.37}    \\
& 2  &  & \checkmark        & 3.76 & 0.63 & 0.73 & 0.57 & 5.74 & 0.44 & 0.59 & 0.32    \\
& 3  & \checkmark & \checkmark & \textbf{3.69}      & \textbf{0.65}       & 0.72       & \textbf{0.59}      & \textbf{5.36}       & \textbf{0.48}       & \textbf{0.61}   & \textbf{0.37} \\
\Xhline{2\arrayrulewidth}          
\end{tabular}
\end{table*}

\section{Dataset and Implementations}
\textbf{Room-to-Room dataset.}
We use the Room-to-Room (R2R) dataset~\cite{anderson2018vision} for evaluating our proposed approach. 
The R2R dataset is 
built upon the Matterport3D dataset~\cite{Matterport3D}.
It consists of 10,800 panoramic views from 194,400 RGB-D images in 90 buildings and has 7,189 paths sampled from its navigation graphs. 
Each path has three ground-truth navigation instructions written by humans. 
The whole dataset has 90 scenes: 61 for training and validation seen, 11 for validation unseen, 18 for test unseen.

\textbf{Evaluation metrics.}
 To compare to existing work, we show the same evaluation metrics used in those works: (1) Navigation Error (NE), mean of the shortest path distance in meters between the agent's final position and the goal location.
(2) Success Rate (SR), the percentage of final positions less than 3m away from the goal location.
(3) Oracle Success Rate (OSR), the success rate if the agent can stop at the closest point to the goal along its trajectory. 
However, we note the importance of a recently added metric that emphasizes the trade-off between success rate and trajectory length: Success rate weighted by (normalized inverse) Path Length (SPL)~\cite{anderson2018evaluation},
which incorporates trajectory lengths and is an important consideration for real-world applications such as robotics.

\textbf{Implementation Details.}
For fair comparison with existing work, we use the pre-trained ResNet-152 on ImageNet to extract image features. 
Following the Self-Monitoring~\cite{ma2019selfmonitoring} and Speaker-Follower~\cite{fried2018speaker} works, the embedded feature vector for each navigable direction is obtained by concatenating an appearance feature with a 4-d orientation feature $[sin \phi; cos \phi; sin \theta; cos \theta]$, where $\phi$ and $\theta$ are the heading and elevation angles.
Please refer to the Appendix for further implementation details.

\section{Evaluation}
\subsection{Comparison with Prior Art}
We first compare the proposed regretful navigation agent with the state-of-the-art methods~\cite{ma2019selfmonitoring, fried2018speaker,wang2019reinforced}. 
As shown in Table~\ref{table:SOTAs}, our method achieves significant performance improvement over the existing approaches.
We achieved 37\% SPL and 48\% SR on the validation unseen set and outperformed all existing work.
Our best performing model achieves 41\% SPL and 50\% SR on validation unseen set when trained with the synthetic data from the Speaker~\cite{fried2018speaker}. 
We demonstrate absolute 8\% SPL improvement and 5\% SR improvement on the test server over the current state-of-the-art method.
We can also see that our regretful navigation agent without data augmentation has already outperformed the existing work on both SR and SPL metrics. 

\subsection{Ablation Study}
Table \ref{table:ablation} shows an ablation study to analyze the effect of each component. The first thing to note is that our method is significantly better than the Self-Monitoring agent which uses greedy decoding, even though it still has a progress monitor loss (although the progress monitor is not used for action selection). 
A second interesting point is that when the Progress Marker is available with the features of each navigable direction that have been visited before, but the Regret Module is not available, performance does not increase significantly (44\% SR). 
Note that we also conducted an experiment with another condition, where the progress monitor estimates were attached to the \textit{forward} embedding, meaning that the network could use that information to improve action selection. 
That condition again was only able to achieve modest gains (45\% SR), compared to our Regret Module which was able to achieve 47\% SR (and 48\% when the Progress Marker was added). 
In all, this shows that the key improvement stems from the design of the Regret Module, allowing the agent to intelligently backtrack after making mistakes.

\begin{table*}[t]
\centering
\small
\renewcommand{\arraystretch}{1.2}
\caption{
    Sanity check for verifying that the source of performance improvement is from the agent's ability to decide when to roll back.
}
\label{table:sanity_rollback}
\begin{tabular}{cccccccccc}
\Xhline{2\arrayrulewidth}
&  Blocking                 & \multicolumn{4}{c}{Validation-Seen} & \multicolumn{4}{c}{Validation-Unseen} \\ \cline{3-10} 
Method  & Rollback       & NE $\downarrow$         & SR $\uparrow$ & OSR $\uparrow$ & SPL  $\uparrow$      & NE $\downarrow$          & SR $\uparrow$ & OSR $\uparrow$    & SPL  $\uparrow$       \\ \hline
\multirow{2}{*}{Self-Monitoring~\cite{ma2019selfmonitoring}}& & 3.72  & 0.63       & \textbf{0.75}  & 0.56    & 5.98   & 0.44       & 0.58   & 0.30 \\ 
&\checkmark & 3.85  & 0.64       & \textbf{0.75}  & 0.58    & 6.02   & 0.44       & 0.60   & 0.34   \\ \hline
\multirow{2}{*}{Regretful} &   & \textbf{3.69} & \textbf{0.65} & 0.72    & 0.59 & \textbf{5.36} & \textbf{0.48} & \textbf{0.61} & 0.37      \\
&\checkmark   & 3.91 & 0.64 & 0.68    & \textbf{0.60} & 5.80 & 0.46 & 0.55 & \textbf{0.41}      \\
\Xhline{2\arrayrulewidth}          
\end{tabular}
\end{table*}

\subsection{Rollback Analysis}

We now further analyze the behavior of the agent to verify that the source of improvement is indeed from the ability to learn when to roll back.

\textbf{Does rollback lead to the performance improvement?}
Our proposed regretful agent relies on the ability to regret and roll back to a previous location,  further exploring the unknown environment to increase the success rate. 
As a sanity check, we manually block all actions leading to rollback for both the state-of-the-art Self-Monitoring agent and our regretful agent\footnote{except when there is only one navigable direction to go.}. 
The result is shown in Table~\ref{table:sanity_rollback}. 
As can be seen, blocking rollback for the Self-Monitoring agent produces mixed results, with worse NE but better metrics such as OSR. The SR, however, is unchanged. 
On the other hand, blocking rollback for our agent significantly reduces most metrics including NE, SR, and OSR especially on unseen environments. 
This shows that blocking the ability to learn when to roll back degrades a large source of performance increase, and this is especially true for unseen environments. 

\begin{figure}[t]
\begin{center}
    \includegraphics[width=0.85\linewidth]{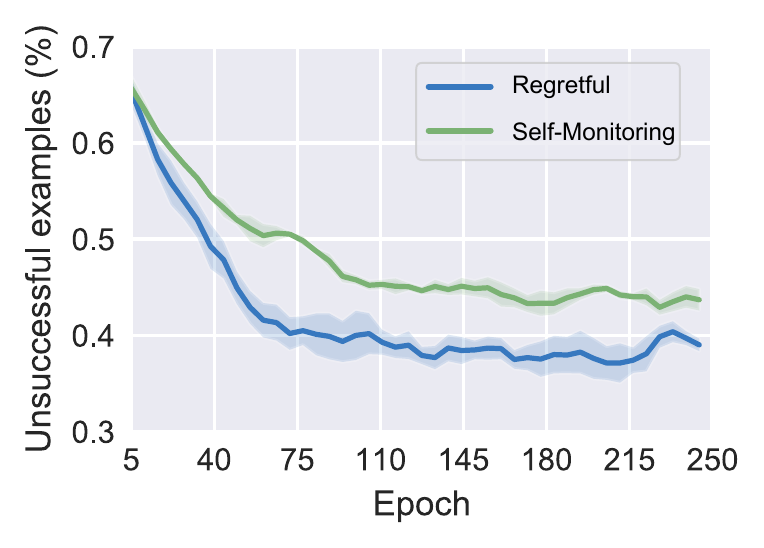}
\end{center}
\vspace{-0.1in}
\caption{
Percentage of unsuccessful examples involving rollback reduced by our proposed regretful agent.
}
\label{fig:rollback_failed}
\vspace{-0.1in}
\end{figure}

\textbf{Number of unsuccessful examples reduced.}
We calculate the total number of unsuccessful examples involves rollback action for both Self-Monitoring and our proposed agent (in percentage). 
As demonstrated in Figure~\ref{fig:rollback_failed}, our proposed regretful agent significantly reduces the unsuccessful examples from around 43\% to 38\%, which correlates to the 4-5\% improvement on SR in Table~\ref{table:SOTAs} and \ref{table:ablation}.

\textbf{Regretful agent in unfamiliar environments.}
The key to the performance increase of an agent focusing on the rollback ability is not that the agent learns a better textual or visual grounding, but that the agent learns to search especially when it is not certain which direction to go. 
To demonstrate this, we train both the Self-Monitoring agent and our proposed regretful agent only on synthetic data and test them on the unseen validation set (real data). 
We expect the regretful agent to outperformed the Self-Monitoring agent across all metrics since our agent is designed to operate in an environment where the agent is likely to be uncertain on action selection. 
As shown in Table~\ref{table:synthetic_only}, when trained using only the synthetic data, our method significantly outperformed Self-Monitoring agent. 
Interestingly, when compared with the Self-Monitoring agent trained with real data, our agent trained with synthetic data is slightly better on ONE, same on OSR, and marginally lower on SR. 
We achieved slightly better performance on oracle metrics since stopping at the correct location is not a hard constrain.
This indicates that even though our regretful agent is not yet learned how to properly stop at the goal (due to training on synthetic data only), the chance that it passes/reaches the goal is slightly higher than Self-Monitoring agent trained with real data.
Further, when the regretful agent trained with real data, the performance improved across all metrics.

\begin{table}[t]
\centering
\scriptsize
\renewcommand{\arraystretch}{1.2}
\caption{
    Ablation study when trained using only the synthetic or real training data.
    Oracle Navigation Error (ONE): the navigation error if the agent can stop at the closest point to the goal along its trajectory.
}
\label{table:synthetic_only}
\begin{tabular}{cccccc}
\Xhline{2\arrayrulewidth}
& & & \multicolumn{3}{c}{Validation-Unseen} \\ \cline{4-6} 
Method  & Synthetic & Real      & ONE $\downarrow$ & SR $\uparrow$ & OSR $\uparrow$    \\ \hline
\multirow{2}{*}{Self-Monitoring~\cite{ma2019selfmonitoring}}&\checkmark &  & 4.09 & 0.35 & 0.49       \\ 
& &\checkmark    & 3.62     & 0.44       & 0.58\\ \hline
\multirow{2}{*}{Regretful}&\checkmark & & 3.47 & 0.41 & 0.58\\
& &\checkmark & \textbf{3.34} & \textbf{0.48} & \textbf{0.61}\\
\Xhline{2\arrayrulewidth}          
\end{tabular}
\end{table}

\subsection{Qualitative Results}
\begin{figure*}[t]
\begin{center}
    \includegraphics[width=1\textwidth]{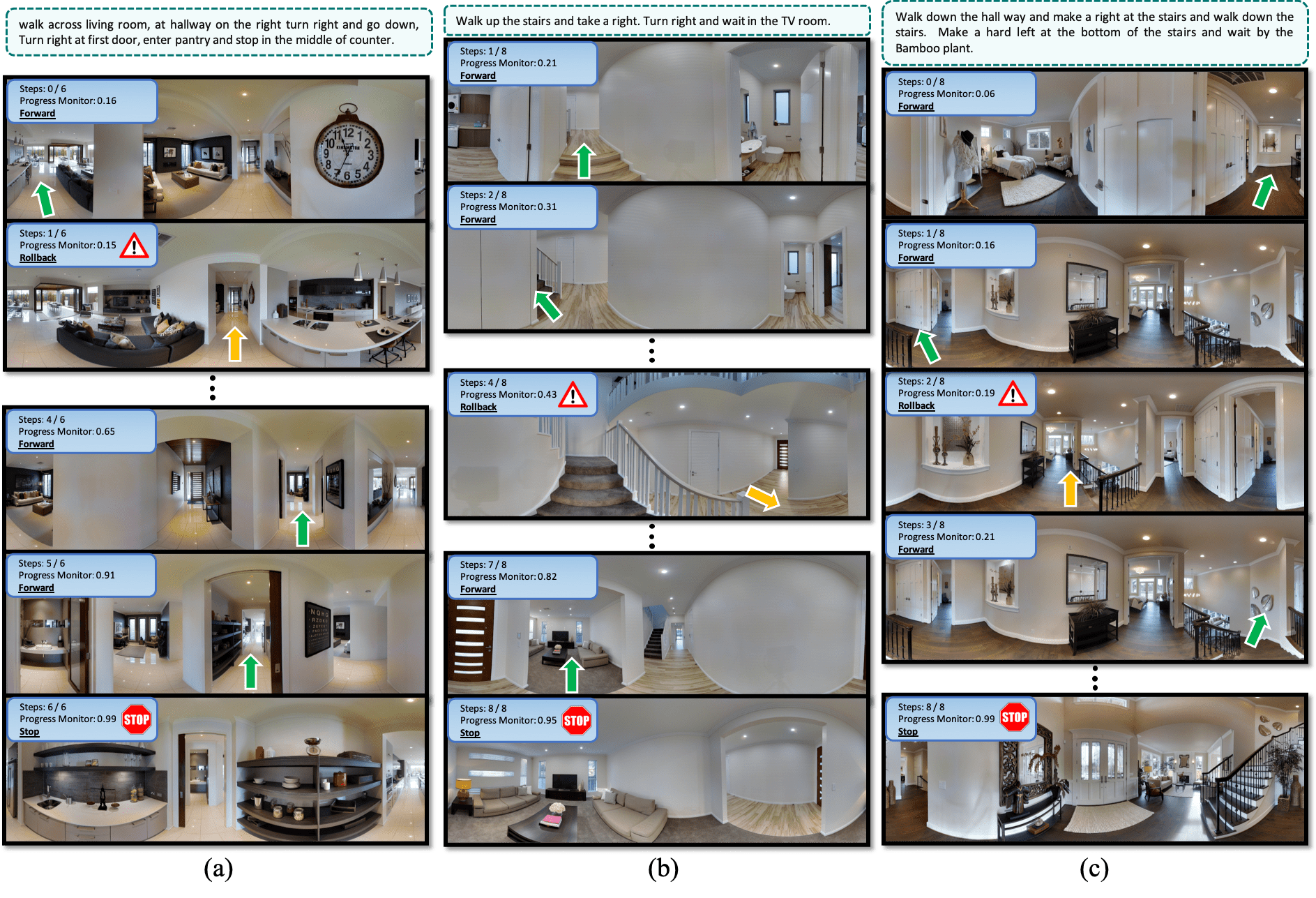}
\end{center}
\vspace{-0.2in}
\caption{
Successful regretful agent navigates in unseen environments.
(a) The agent made a mistake at the first step, but it was able to roll back to the previous location since the output of the progress monitor was not significantly increased. It then follows the rest of the instruction correctly.
(b) The agent is able to correctly follow the instruction at the beginning but made a mistake by walking up the stairs again.
The agent realized that the output of the progress monitor is decreased and the next action \textit{take a right} is not feasible and decides to rollback rollback at step 4. 
The agent was then able to follow the rest of the instruction and stop with estimated progress 0.95.
(c) The agent made a mistake by missing the stairs at step 1. 
It was however able to decide to rollback at step 2 and moves down stairs as instructed and successfully stops near the bamboo plant with estimated progress 0.99.
Please see Appendix for the full trajectories.
}
\label{fig:unseen_success}
\vspace{-0.1in}
\end{figure*}

Figures \ref{fig:unseen_success} show qualitative outputs of our model during successful navigation in unseen environments. 
In Figure \ref{fig:unseen_success} (a), the agent made a mistake at the first step, and the estimated progress at the second step slightly decreases.
The agent then decides to rollback, after which the progress monitor significantly increases. 
Finally, the agent stopped correctly as instructed. 
Figure \ref{fig:unseen_success} (b) shows an example where the agent correctly goes up the stairs but incorrectly does it again rather than turning and finding the TV as instructed. 
Note that the progress monitor increases but only by a small amount; this demonstrates the need for learned mechanisms that can reason about the textual and visual grounding and context, as well as the resulting level of change in progress. 
In this case the agent then correctly decides to rollback and successfully walked into the TV room. 
Similarly, in Figure \ref{fig:unseen_success} (c), the agent misses the stairs, resulting in a very small progress increase. 
The agent decides to rollback as a result. 
Upon reaching the goal, the agent's progress estimate is 99\%.
Please refer to the Appendix for the full trajectories and unsuccessful examples.

\section{Conclusion}
In this paper, we have proposed a end-to-end trainable regretful navigation agent for the VLN task. Inspired by the intuition of viewing this task as graph search over the navigation graph, we specifically use a progress monitor as a learned heuristic that can be trained and employed during inference to greedily select the next best action (best-first search). The progress monitor incorporates information from grounded language instructions and visual information, integrated across time with LSTMs. We then propose a Regret Module that is able to learn to decide when to perform backtracking depending on the progress made and state of the agent. 
Finally, a Progress Marker is used to allow the agent to reason about previous visits and unvisited directions, so that the agent can choose a better navigable direction by reducing action probabilities for visited locations with lower progress estimate. 

The resulting framework is able to achieve state-of-the-art success rates compared to existing published methods on the public leaderboard, without using beam search. 
We show through several extensive analyses that the source of performance improvement is the design of the learned rollback mechanism, that when blocked the performance decreases, and that this learning can occur even on purely synthetic data and generalize to real data.
We also demonstrated that the total number of unsuccessful examples involve rollback reduces with our regretful agent.
There is a great deal of future work possible, which extends our method. For example, other aspects of graph search can be incorporated such as elements of exploration (e.g. a search space frontier), but in a manner that is more efficient than beam search, can also be investigated. Finally, a combination of goal-driven perception and reinforcement learning would be interesting to explore, as the tasks contain less and less structured information (e.g. embodied QA).

\section*{Acknowledgments}
This research was partially supported by DARPA’s Lifelong Learning Machines (L2M) program, under Cooperative Agreement HR0011-18-2-001.
We thank Chia-Jung Hsu for her valuable and artistic suggestions on the figures.
\section*{Appendix}
\appendix
\section{Network Architecture}
The embedding dimension of the instruction encoder is 256, followed by a dropout layer with ratio 0.5.
We encode the instruction using a regular LSTM, and the hidden state is 512 dimensional.
The MLP $g$ used for projecting the raw image feature is ${BN \xrightarrow{} FC \xrightarrow{} BN \xrightarrow{} Dropout \xrightarrow{} ReLU}$. 
The FC layer projects the 2176-d input vector to a 1024-d vector, and the dropout ratio is set to be 0.5.
The hidden state of the LSTM which allows integration of information across time is 512.
When using the progress marker, the markers are tiled $n=32$ times. 
The dimension of the learnable matrices are:
$\mW_{x} \in \mathbb{R}^{512 \times 512}$, 
$\mW_{v} \in \mathbb{R}^{512 \times 1024}$, 
$\mW_{a} \in \mathbb{R}^{1024 \times 1024}$, 
$\mW_{r} \in \mathbb{R}^{1 \times 2}$, 
$\mW_{fr} \in \mathbb{R}^{1024 \times 1024}$ without progress marker,
and $\mW_{fr} \in \mathbb{R}^{1024 \times 1056}$ with progress marker.

\section{Comparison with Beam Search Methods}
We compare our method using greedy action selection with existing beam search approaches, e.g., Pragmatic Inference in Speaker-Follower~\cite{fried2018speaker} and progressed integrated beam search in Self-Monitoring agent~\cite{ma2019selfmonitoring}.
We can see in Table~\ref{table:greedy-vs-beam} that, while beam search methods perform well on success rate (SR), their trajectory lengths are significantly longer, achieving low success rate weighted by Path Length (SPL) scores and therefore are impractical for real-world applications.
On the other hand, our proposed method significantly improved both SR and SPL when not using beam search.
\begin{table}[t]
\vspace{-0.1in}
\centering
\scriptsize
\renewcommand{\arraystretch}{1.2}
\caption{
Comparison of our regretful agent using greedy action selection with beam search.
}
\label{table:greedy-vs-beam}
\begin{tabular}{cccccc}
\Xhline{2\arrayrulewidth}
& Beam & \multicolumn{4}{c}{Test set (leaderboard)} \\ \cline{3-6} 
Method  & search      & NE $\downarrow$ & SR $\uparrow$ & Length $\downarrow$ & SPL $\uparrow$   \\ \hline
\multirow{2}{*}{Speaker-Follower~\cite{fried2018speaker}} &  & 6.62 & 0.35  & 14.82  & 0.28 \\
& $\triangle$ & 4.87 & 0.53  & 1257.38 & 0.01 \\ \hline
\multirow{2}{*}{Self-Monitoring~\cite{ma2019selfmonitoring}}&  & 5.99 & 0.43 & 17.11 & 0.32        \\ 
& $\triangle$    & \textbf{4.48}     & \textbf{0.61} & 373.09       & 0.02 \\ \hline
Regretful & & 5.69 & 0.48 & \textbf{13.69} & \textbf{0.40}\\
\Xhline{2\arrayrulewidth}          
\end{tabular}
\vspace{-0.1in}
\end{table}

\section{Qualitative Analysis}
\subsection{Successful examples}
\begin{figure*}[t]
\begin{center}
    \includegraphics[width=0.8\textwidth]{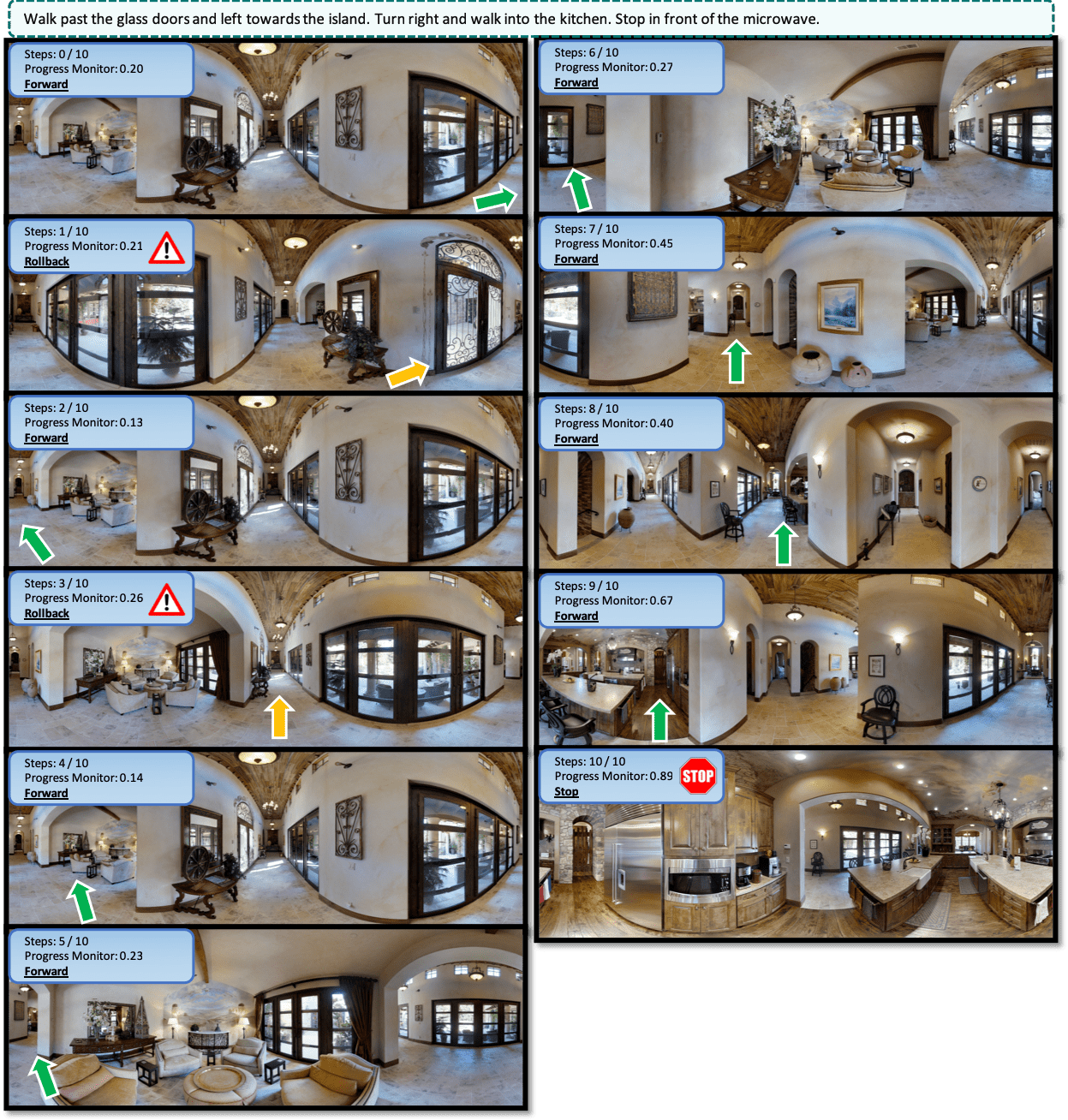}
\end{center}
\caption{
The first part of the instruction \textit{walk past the glass doors} is ambiguous since there are multiple directions that lead to glass doors, and naturally the agent is confused and uncertain where to go. 
Our agent is able to perform local search on the navigation graph and decides to roll back multiple times at the beginning of the navigation.
At step 6, the agent performs an action \textit{turn right}.
Consequently, the progress estimate at step 7 significantly increased to 45\%. 
Interestingly, the agent continues to move forward even though the progress estimate slightly decreased from step 7 to step 8. 
We reckon that this as one of the advantage of using a  learning-based regret module as opposed to using a  hard-coded threshold.
The agent then successfully follows the instruction and \textit{stops in front of the microwave} with progress estimate 89\%.
}
\label{fig:unseen_success_4}
\end{figure*}

\begin{figure}[t]
\begin{center}
    \includegraphics[width=0.4\textwidth]{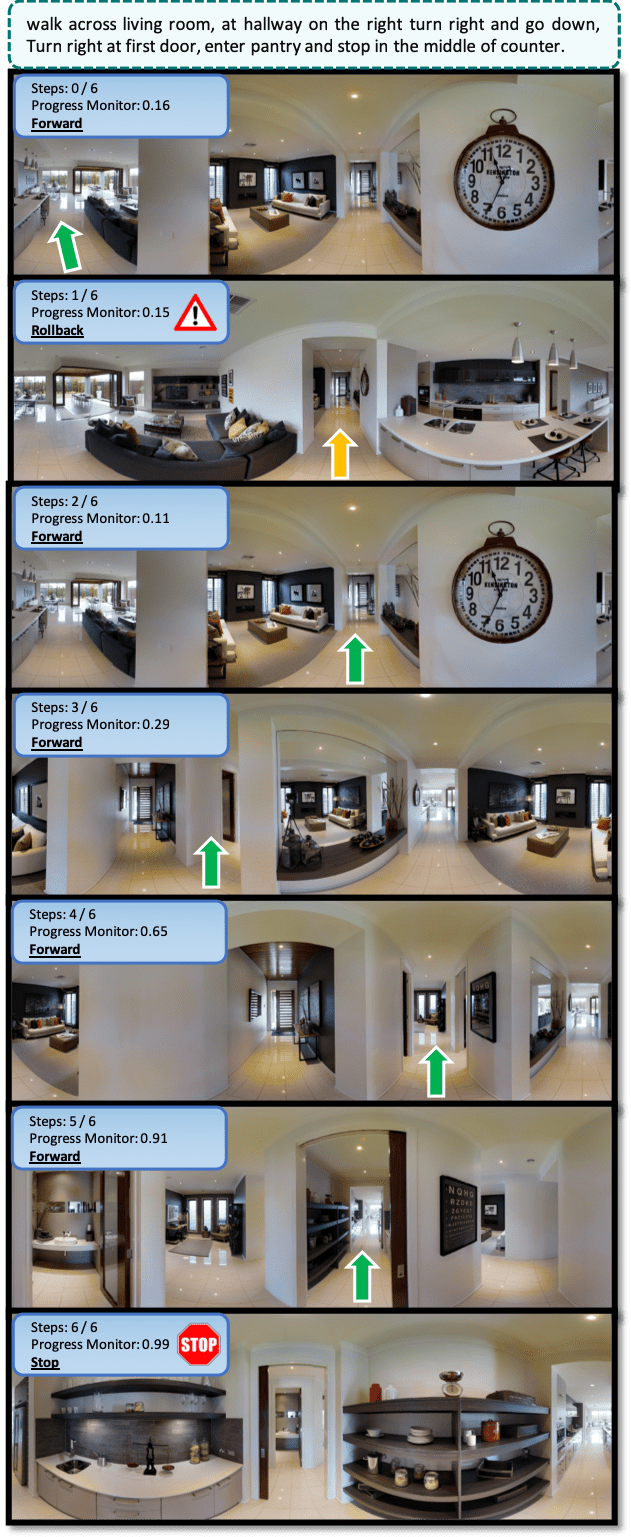}
\end{center}
\caption{
The agent first \textit{walk across living room}, but decides to move into the direction that leads to kitchen and dinning room. 
At step 1, the agent decides to roll back due to a decreasing of the progress monitor output. 
The agent then followed the rest of the instruction successfully with the progress monitor steadily increased at each step. 
Finally, the agent decides to stop with the progress estimate 99\%. 
}
\label{fig:unseen_success_1}
\end{figure}

\begin{figure}[t]
\begin{center}
    \includegraphics[width=0.33\textwidth]{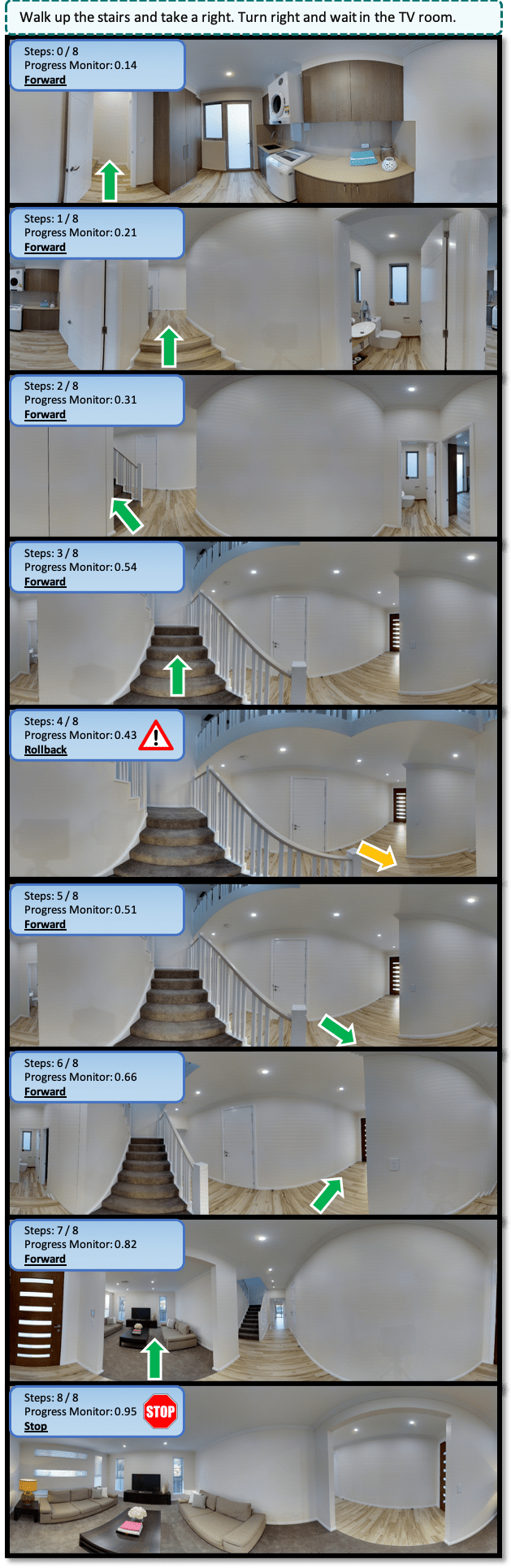}
\end{center}
\caption{
The agent walked up the stairs as instructed at step 1, but the second set of stairs makes the instruction ambiguous.
The agent continues to walk up stairs but soon realized that it needs to go down the stairs and \textit{turn right} from step 4 - 6. 
When the agent decides to turn right, we can see the progress estimate significantly increased from 51\% to 66\%. 
As the agent turned right to the TV room, the progress estimate increased again to 82\%. 
Finally, the agent stops with the progress monitor output 95\%. 
}
\label{fig:unseen_success_2}
\vspace{-0.1in}
\end{figure}

\begin{figure}[t]
\begin{center}
    \includegraphics[width=0.32\textwidth]{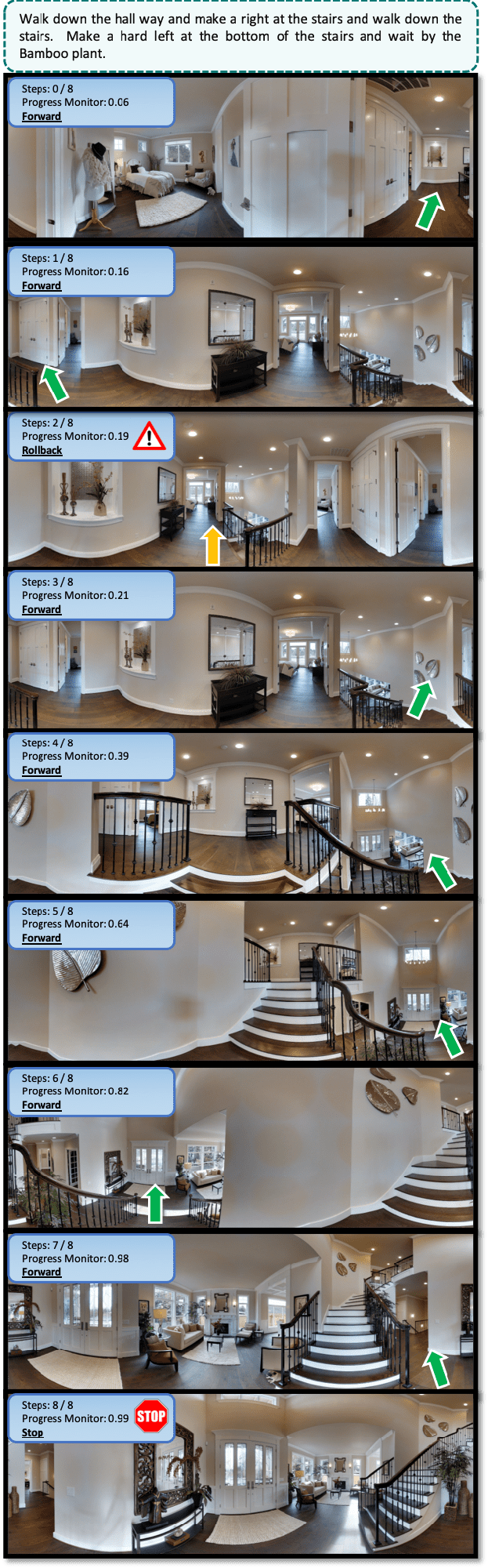}
\end{center}
\caption{
The agent walks down the hall way to the stairs but failed to \textit{walk down the stairs} at step 1.
With a small increase on the progress monitor output, the agent then decides to roll back and take the action to walk down the stairs. 
Once walking down, we can see the progress estimate increased to 39\%, and as the agent goes further down, the progress estimate reached 98\% at the bottom of the stairs. 
Finally, the agent decides to stop near by the bamboo plant with progress estimate 99\%. 
}
\label{fig:unseen_success_3}
\end{figure}

We show the complete trajectory of the agents successfully deciding when to roll back and reach the goal in unseen environments in Figure~\ref{fig:unseen_success_4}, \ref{fig:unseen_success_1}, \ref{fig:unseen_success_2}, and \ref{fig:unseen_success_3}. 

In Figure~\ref{fig:unseen_success_4}, we demonstrate that the agent is capable of performing a local search on the navigation graph.
Specifically, from step 0 to step 3, the agent searched two possible directions and decided to move with one particular direction at step 4.
Once it reached step 5, the agent decides to continue to move forward, and we observed that the progress estimate significantly increased to 45\% at step 7. 
Interestingly, unlike other examples we have shown, the agent did not decide to roll back despite the progress estimate slightly decreased from 45\% to 40\%. 
We reckon that this is one of the advantages of using a learning-based regret module, where a learned and dynamically changing threshold decides when to rollback.
Finally, the agent successfully stopped in front of the microwave.

In Figure~\ref{fig:unseen_success_1}, the agent is instructed to \textit{walk across living room}.
It is ambiguous since both directions seem like a living room. 
Our agent first decides to move into the direction that leads to a room with a kitchen and living room. 
It then decided to roll back with the progress monitor output slightly decreased. 
The agent then followed the rest of the instruction successfully with the progress monitor steadily increased at each step after that. 
Finally, the agent decides to stop with the progress estimate 99\%.

In Figure~\ref{fig:unseen_success_2}, the agent first moved out of the room and walked up the stairs as instructed, but the second set of stairs makes the instruction ambiguous.
The agent continued to walk up the stairs for one more step and then decided to go down the stairs at step 4.
As the agent decided to turn right at step 6, we can see the progress estimate significantly increased from 51\% to 66\%. 
Once the agent entered the TV room, the progress estimate increased again to 82\%. 
Finally, the agent successfully stopped with the progress monitor output 95\%. 

In Figure~\ref{fig:unseen_success_3}, the agent failed to \textit{walk down the stairs} at step 1.
Because of the proposed Regret Module and Progress Marker, the agent was able to discover the correct path to go downstairs. 
Once walking down, the progress estimate increased to 39\% immediately, and as the agent goes further down, the progress estimate reached 98\% by the time the agent reached the bottom of the stairs. 
Finally, the agent decided to wait by the bamboo plant with progress estimate 99\%. 

\subsection{Failed examples}
We have shown how the agent can successfully utilize the rollback mechanism to reach the goal, even though it is not familiar with the environment and likely to be uncertain about some actions it took.
Intuitively, the rollback mechanism can increase the chance that the agent reaches the goal as long as the agent can correctly decide when to stop. 

We now discuss two failed examples of our proposed regretful agent in unseen environments that highly resemble the successful examples in terms of the given instruction and ground-truth path.
Both examples demonstrate that the agent successfully rolled back to the correct path towards the goal but failed to stop at the goal.

\begin{figure}[t]
\begin{center}
    \includegraphics[width=0.33\textwidth]{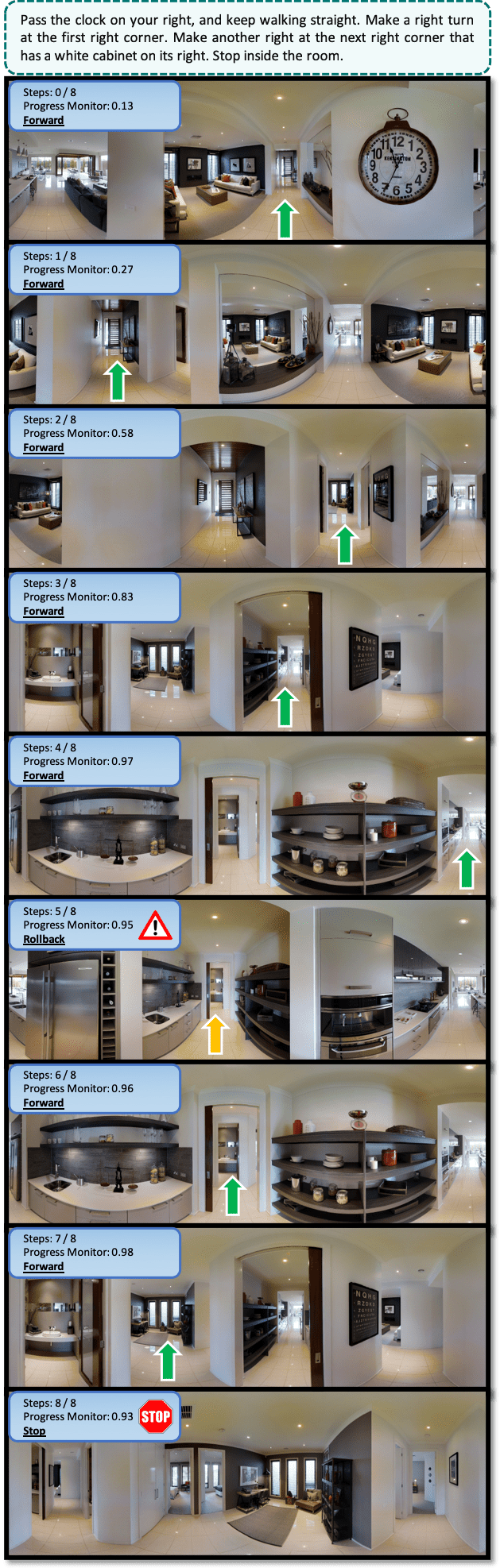}
\end{center}
\caption{
\textbf{Failed example.}
The agent starts to navigate through the unseen environment by following the given instruction.
It was able to successfully follow the instruction and correctly reach the goal at step 4. 
The agent then decided to move forward towards the kitchen and correctly decided to roll back to the goal. 
However, the agent did not stop and continue to explore the environment and eventually stopped a bit further from the goal.
}
\label{fig:unseen_failed_1}
\end{figure}

Specifically, in Figure~\ref{fig:unseen_failed_1}, the agent reaches the room with the white cabinet as instructed but decided to move one step forward. 
The agent then decided to roll back to the room correctly at step 5. 
However, this does not help the agent to stop at the goal resulting in a failed run. 

\begin{figure*}[t]
\begin{center}
    \includegraphics[width=0.8\textwidth]{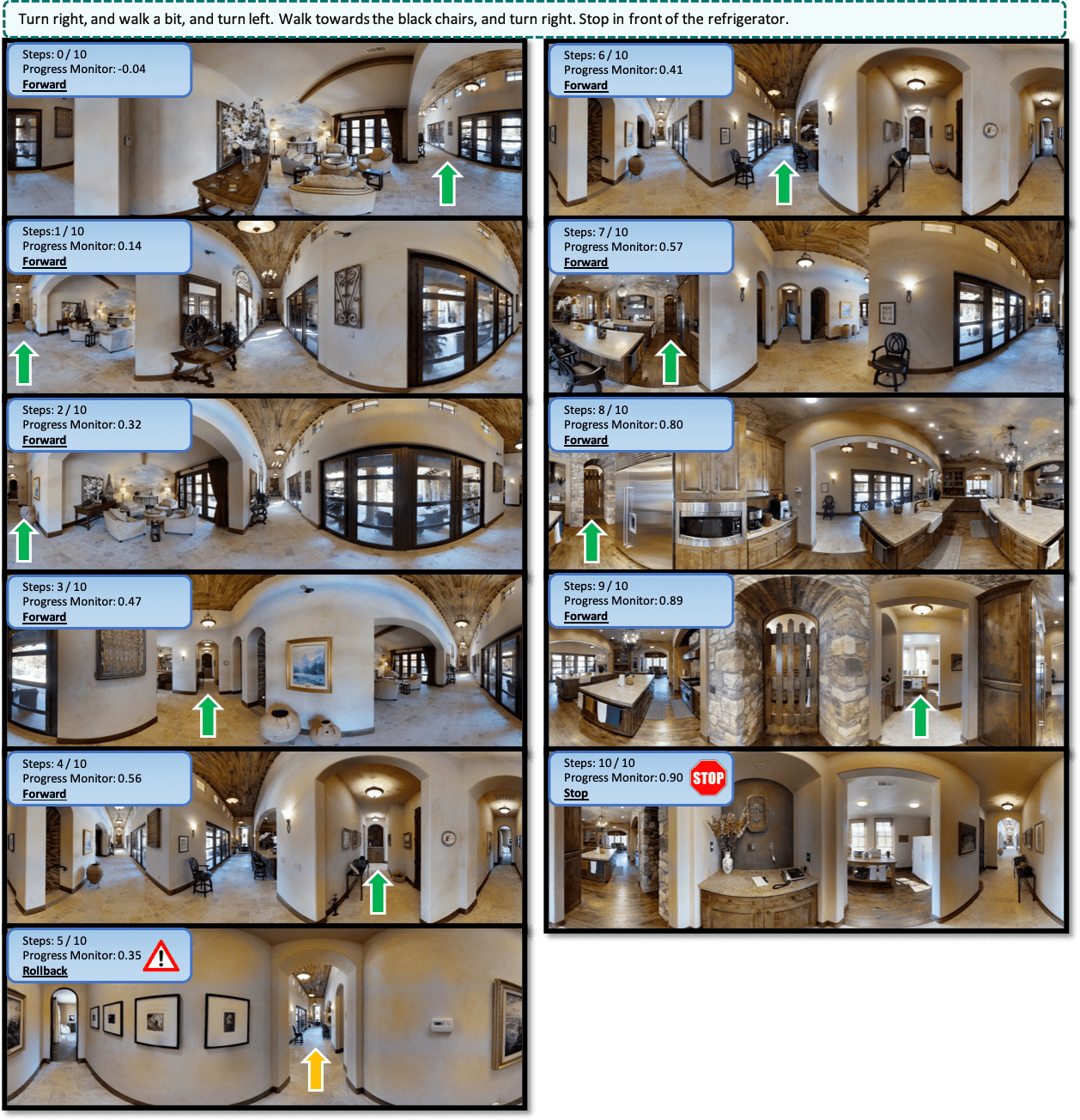}
\end{center}
\caption{
The agent correctly followed the first parts of the instruction until step 4, but it decided to move forward towards the hall. 
At step 5, the agent correctly decided to roll back with the progress estimate decreased from 56\% to 35\%.
The agent was then able to follow the rest of the instruction successfully and reach the refrigerator at step 8. 
However, the agent did not stop nearby the refrigerator and continued to take another two forward steps.
}
\label{fig:unseen_failed_2}
\end{figure*}

On the other hand, in Figure~\ref{fig:unseen_failed_2}, we can see that the progress estimate at step 5 significantly dropped by 21\%, and the agent correctly decided to roll back.
The agent then successfully reached the refrigerator but did not stop immediately. 
It continued to move forward after step 8, resulting in an unsuccessful run.

\begin{figure*}[t]
\begin{center}
    \includegraphics[width=0.8\textwidth]{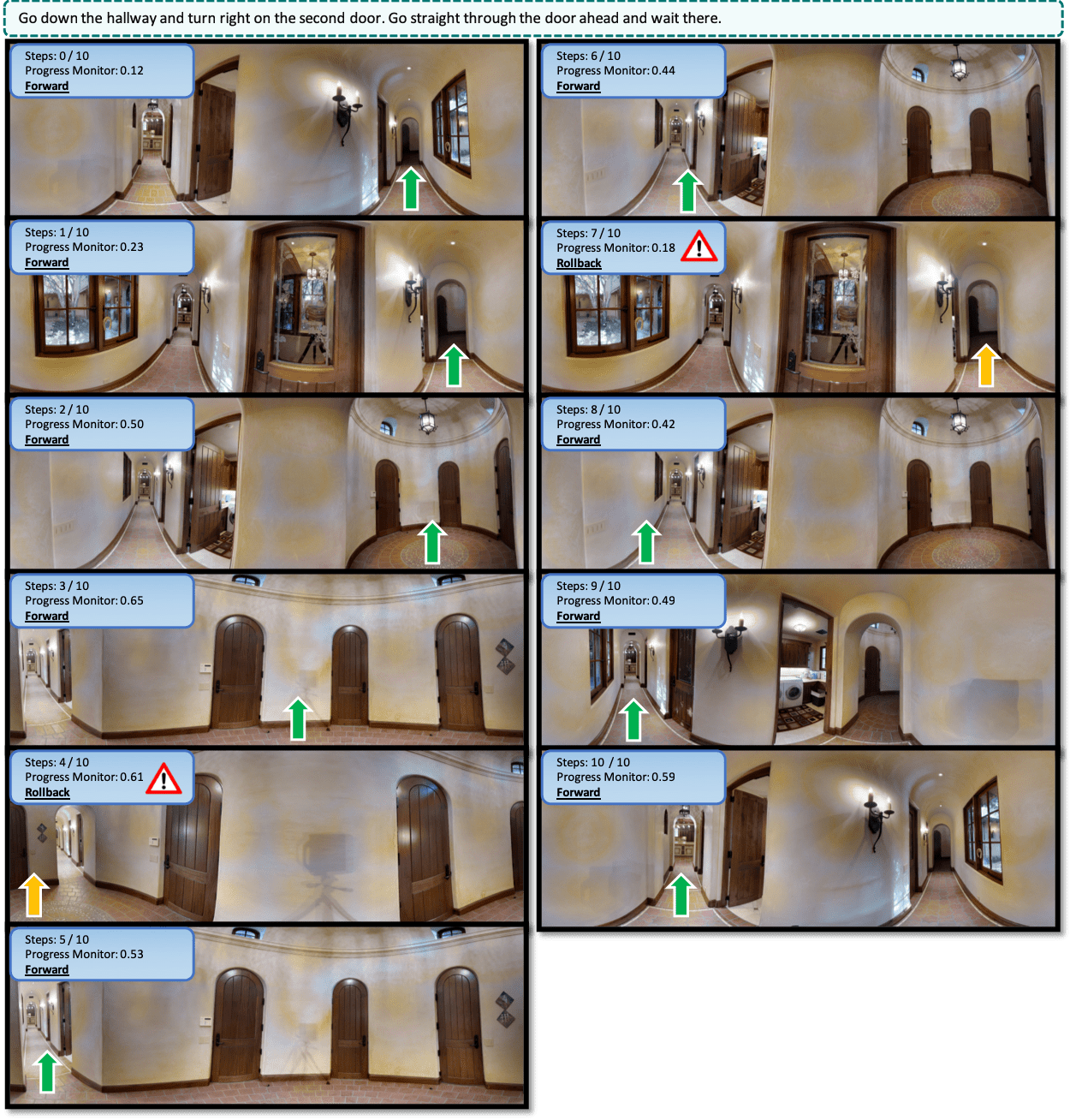}
\end{center}
\caption{
The agent followed the first part of instruction to \textit{go down the hallway}.
As the agent reached the end of the hallway, it was not able to find the second door to turn left. 
The agent then decided to roll back at step 4 with progress estimate decreased from 65\% to 61\%. 
The agent continued to go back towards the hallway but decided to roll back again at step 7. 
Although the agent was able to correct its errors made at the first few steps and \textit{escape} from the hallway leading to the dead end, it ends up unsuccessful.
}
\label{fig:unseen_failed_3}
\end{figure*}

Lastly, we discuss a failed example when the agent incorrectly decided when to roll back.
In Figure~\ref{fig:unseen_failed_3}, the agent first followed the instruction to \textit{go down the hallway} and tried to find the second door to turn right. 
As the agent reached the end of the hallway at step 4, it decided to roll back since there is no available navigable direction that leads to \textit{turn right}. 
The agent then decided to go down the hallway again with completely opposite direction.
However, the agent decided to roll back again at step 7 with the progress estimate dropped to 18\%. 
Although the agent eventually was able to \textit{escape} from the hallway leading to the dead end, it ends up unsuccessful.

\clearpage{\thispagestyle{empty}\cleardoublepage}
{\small
\bibliographystyle{ieee}
\bibliography{egbib}

\begin{thebibliography}{10}\itemsep=-1pt

\bibitem{anderson2018evaluation}
P.~Anderson, A.~Chang, D.~S. Chaplot, A.~Dosovitskiy, S.~Gupta, V.~Koltun,
  J.~Kosecka, J.~Malik, R.~Mottaghi, M.~Savva, et~al.
\newblock On evaluation of embodied navigation agents.
\newblock {\em arXiv preprint arXiv:1807.06757}, 2018.

\bibitem{anderson2018vision}
P.~Anderson, Q.~Wu, D.~Teney, J.~Bruce, M.~Johnson, N.~S{\"u}nderhauf, I.~Reid,
  S.~Gould, and A.~van~den Hengel.
\newblock Vision-and-language navigation: Interpreting visually-grounded
  navigation instructions in real environments.
\newblock In {\em Proceedings of the IEEE Conference on Computer Vision and
  Pattern Recognition (CVPR)}, volume~2, 2018.

\bibitem{barraquand1991robot}
J.~Barraquand and J.-C. Latombe.
\newblock Robot motion planning: A distributed representation approach.
\newblock {\em The International Journal of Robotics Research}, 10(6):628--649,
  1991.

\bibitem{bhardwaj2017learning}
M.~Bhardwaj, S.~Choudhury, and S.~Scherer.
\newblock Learning heuristic search via imitation.
\newblock {\em arXiv preprint arXiv:1707.03034}, 2017.

\bibitem{brodeur2017home}
S.~Brodeur, E.~Perez, A.~Anand, F.~Golemo, L.~Celotti, F.~Strub, J.~Rouat,
  H.~Larochelle, and A.~Courville.
\newblock Home: A household multimodal environment.
\newblock {\em arXiv preprint arXiv:1711.11017}, 2017.

\bibitem{Matterport3D}
A.~Chang, A.~Dai, T.~Funkhouser, M.~Halber, M.~Niessner, M.~Savva, S.~Song,
  A.~Zeng, and Y.~Zhang.
\newblock {Matterport3D}: Learning from {RGB-D} data in indoor environments.
\newblock {\em International Conference on 3D Vision (3DV)}, 2017.

\bibitem{das2018embodied}
A.~Das, S.~Datta, G.~Gkioxari, S.~Lee, D.~Parikh, and D.~Batra.
\newblock Embodied question answering.
\newblock In {\em Proceedings of the IEEE Conference on Computer Vision and
  Pattern Recognition (CVPR)}, 2018.

\bibitem{das2018neural}
A.~Das, G.~Gkioxari, S.~Lee, D.~Parikh, and D.~Batra.
\newblock {N}eural {M}odular {C}ontrol for {E}mbodied {Q}uestion {A}nswering.
\newblock In {\em Proceedings of the Conference on Robot Learning (CoRL)},
  2018.

\bibitem{de2018talk}
H.~de~Vries, K.~Shuster, D.~Batra, D.~Parikh, J.~Weston, and D.~Kiela.
\newblock Talk the walk: Navigating new york city through grounded dialogue.
\newblock {\em arXiv preprint arXiv:1807.03367}, 2018.

\bibitem{eysenbach2018leave}
B.~Eysenbach, S.~Gu, J.~Ibarz, and S.~Levine.
\newblock Leave no trace: Learning to reset for safe and autonomous
  reinforcement learning.
\newblock In {\em Proceedings of the International Conference on Learning
  Representations (ICLR)}, 2018.

\bibitem{fikes1972learning}
R.~E. Fikes, P.~E. Hart, and N.~J. Nilsson.
\newblock Learning and executing generalized robot plans.
\newblock {\em Artificial intelligence}, 3:251--288, 1972.

\bibitem{fried2018speaker}
D.~Fried, R.~Hu, V.~Cirik, A.~Rohrbach, J.~Andreas, L.-P. Morency,
  T.~Berg-Kirkpatrick, K.~Saenko, D.~Klein, and T.~Darrell.
\newblock Speaker-follower models for vision-and-language navigation.
\newblock In {\em Advances in Neural Information Processing Systems (NIPS)},
  2018.

\bibitem{ai2thor}
E.~Kolve, R.~Mottaghi, D.~Gordon, Y.~Zhu, A.~Gupta, and A.~Farhadi.
\newblock {AI2-THOR: An Interactive 3D Environment for Visual AI}.
\newblock {\em arXiv}, 2017.

\bibitem{ma2019selfmonitoring}
C.-Y. Ma, J.~Lu, Z.~Wu, G.~AlRegib, Z.~Kira, R.~Socher, and C.~Xiong.
\newblock Self-monitoring navigation agent via auxiliary progress estimation.
\newblock In {\em Proceedings of the International Conference on Learning
  Representations (ICLR)}, 2019.

\bibitem{mirowski2017learning}
P.~Mirowski, R.~Pascanu, F.~Viola, H.~Soyer, A.~J. Ballard, A.~Banino,
  M.~Denil, R.~Goroshin, L.~Sifre, K.~Kavukcuoglu, et~al.
\newblock Learning to navigate in complex environments.
\newblock In {\em Proceedings of the International Conference on Learning
  Representations (ICLR)}, 2017.

\bibitem{mnih2016asynchronous}
V.~Mnih, A.~P. Badia, M.~Mirza, A.~Graves, T.~Lillicrap, T.~Harley, D.~Silver,
  and K.~Kavukcuoglu.
\newblock Asynchronous methods for deep reinforcement learning.
\newblock In {\em International conference on machine learning (ICML)}, pages
  1928--1937, 2016.

\bibitem{mousavian2018visual}
A.~Mousavian, A.~Toshev, M.~Fiser, J.~Kosecka, and J.~Davidson.
\newblock Visual representations for semantic target driven navigation.
\newblock {\em arXiv preprint arXiv:1805.06066}, 2018.

\bibitem{russell2016artificial}
S.~J. Russell and P.~Norvig.
\newblock {\em Artificial intelligence: a modern approach}.
\newblock Malaysia; Pearson Education Limited,, 2016.

\bibitem{savva2017minos}
M.~Savva, A.~X. Chang, A.~Dosovitskiy, T.~Funkhouser, and V.~Koltun.
\newblock Minos: Multimodal indoor simulator for navigation in complex
  environments.
\newblock {\em arXiv preprint arXiv:1712.03931}, 2017.

\bibitem{wang2019reinforced}
X.~Wang, Q.~Huang, A.~Celikyilmaz, J.~Gao, D.~Shen, Y.-F. Wang, W.~Y. Wang, and
  L.~Zhang.
\newblock Reinforced cross-modal matching and self-supervised imitation
  learning for vision-language navigation.
\newblock In {\em Proceedings of the IEEE Conference on Computer Vision and
  Pattern Recognition (CVPR)}, 2019.

\bibitem{wang2018look}
X.~Wang, W.~Xiong, H.~Wang, and W.~Y. Wang.
\newblock Look before you leap: Bridging model-free and model-based
  reinforcement learning for planned-ahead vision-and-language navigation.
\newblock In {\em European Conference on Computer Vision (ECCV)}, 2018.

\bibitem{wayne2018unsupervised}
G.~Wayne, C.-C. Hung, D.~Amos, M.~Mirza, A.~Ahuja, A.~Grabska-Barwinska,
  J.~Rae, P.~Mirowski, J.~Z. Leibo, A.~Santoro, et~al.
\newblock Unsupervised predictive memory in a goal-directed agent.
\newblock {\em arXiv preprint arXiv:1803.10760}, 2018.

\bibitem{wilt2015building}
C.~M. Wilt and W.~Ruml.
\newblock Building a heuristic for greedy search.
\newblock In {\em Eighth Annual Symposium on Combinatorial Search}, 2015.

\bibitem{wu2018building}
Y.~Wu, Y.~Wu, G.~Gkioxari, and Y.~Tian.
\newblock Building generalizable agents with a realistic and rich 3d
  environment.
\newblock {\em arXiv preprint arXiv:1801.02209}, 2018.

\bibitem{xia2018gibson}
F.~Xia, A.~R. Zamir, Z.~He, A.~Sax, J.~Malik, and S.~Savarese.
\newblock Gibson env: Real-world perception for embodied agents.
\newblock In {\em Proceedings of the IEEE Conference on Computer Vision and
  Pattern Recognition}, pages 9068--9079, 2018.

\bibitem{xu2007discriminative}
Y.~Xu, A.~Fern, and S.~W. Yoon.
\newblock Discriminative learning of beam-search heuristics for planning.
\newblock In {\em IJCAI}, pages 2041--2046, 2007.

\bibitem{yu2018guided}
H.~Yu, X.~Lian, H.~Zhang, and W.~Xu.
\newblock Guided feature transformation (gft): A neural language grounding
  module for embodied agents.
\newblock {\em arXiv preprint arXiv:1805.08329}, 2018.

\end{thebibliography}
}

\end{document}